%% file: sample_paper.tex
\documentclass[twoside]{article}

\input{extrapackages}

\usepackage[accepted]{aistats2024}
\usepackage[round]{natbib}

\begin{document}

%

%

\twocolumn[

\aistatstitle{How Good is a Single Basin?}

\aistatsauthor{ Kai Lion \And Lorenzo Noci \And Thomas Hofmann \And Gregor Bachmann}

\aistatsaddress{ ETH Zürich, Dept of Computer Science } ]

\begin{abstract}
The multi-modal nature of neural loss landscapes is often considered to be the main driver behind the empirical success of deep ensembles. In this work, we probe this belief by constructing various "connected" ensembles which are restricted to lie in the same basin. Through our experiments, we demonstrate that increased connectivity indeed negatively impacts performance. However, when incorporating the knowledge from other basins implicitly through distillation, we show that the gap in performance can be mitigated by re-discovering (multi-basin) deep ensembles within a single basin. Thus, we conjecture that while the extra-basin knowledge is at least partially present in any given basin, it cannot be easily harnessed without learning it from other basins.
\end{abstract}

\section{Introduction}
The intricate characteristics of neural loss landscapes pose one of the most intriguing puzzles in deep learning. The surfaces are characterized by their highly non-convex nature, giving rise to an exponential number of minima belonging to various \textit{modes} or \textit{basins}. Simple, first-order optimizers such as stochastic gradient descent are then used to navigate these landscapes, leading to a complex interplay between stochasticity and non-convexity. As a consequence, a plethora of interesting behaviours are observed empirically: (1)  models trained under different random initializations and batch orderings end up in different parts of the weight space while maintaining similar generalization performance \citep{choromanska_loss_2015}, (2) these independently trained models can be connected by simple but non-linear curves \citep{draxler_essentially_2018, garipov_loss_2018} in parameter space, and (3) ensembling these leads to an increase in performance \citep{lakshminarayanan_simple_2017}. The gain in performance can be attributed to the predictive diversity of the obtained models (akin to the classic bias-variance tradeoff), i.e. different runs of the same model result in "different interpretations of the data" due to the non-convexity and the stochasticity of the learning process. The idea of distinguishing between models from different basins is essential and has been used as a synonym for higher model diversity. The common belief is that \textit{diversity is achieved by visiting different basins in the loss landscape}.  This belief suggests that ensembling models from the same basin is of limited use, as models are not sufficiently diverse. Indeed, many works use this interpretation of diversity as a design principle for their inference algorithms \citep{huang_snapshot_2017, dangelo_repulsive_2021, loshchilov_sgdr_2017, zhang_cyclical_2020}.

In this work, we take the opposite direction and investigate instead how performant ensembles can be if they are restricted to lie in a single basin. To that end, we design a set of ensemble baselines that increasingly manage to close the gap to deep ensembles in terms of test accuracy with the additional constraint that each ensemble member lies in the same basin. We coin such ensembles \textit{connected ensembles} alluding to the fact that models from the same basin are linearly mode-connected. Our more intuitive approaches for constructing connected ensembles suggest that increased connectivity is often associated with lower performance. However, by leveraging the insights of \citet{frankle_linear_2020} regarding the stability of SGD and adopting the distillation procedure introduced by \cite{hinton_distilling_2015}, this relationship can be disrupted. Connected ensembles constructed using this distillation method show significantly improved predictive performance. We further demonstrate the broad applicability of the connected ensemble class on two model families and three widely used benchmarks. More specifically, we consider ensembles of \textit{ResNets} \citep{he_deep_2016} and the more recent \textit{Vision Transformers} \citep{dosovitskiy_image_2021}.

These results have strong implications for the design of learning algorithms that leverage multiple model runs, demonstrating that in principle, one does not need to break linear mode-connectivity to find diverse models but can rather ensemble models that are within the same basin. From a theoretical point of view, our results on distilled ensembles imply that the information of multiple basins can be re-discovered within a single basin to a large extent. However, more work is necessary to assess whether this knowledge can be leveraged without the explicit aid of other basins. More specifically, we make the following contributions:
\begin{itemize}
    \item We design a rich set of connected ensembles and characterize a trade-off between diversity and connectivity.
    \item We show that implicitly incorporating knowledge from other basins allows us to design strong connected ensembles that significantly close the gap in performance to deep ensembles. We thus demonstrate that a single basin could suffice for the construction of diverse ensembles.
\end{itemize}

The structure of this paper is as follows. In Section \ref{sec:back}, we introduce the background and motivate the research question. In Section \ref{sec:exploring_single_basin}, we describe methods to explore a single loss basin. We expand on those results by further enhancing the single-basin ensemble performance with different techniques in Section \ref{sec:rediscovering_single_basin}. We provide a review of related work in Section \ref{sec:related-work} and discuss implications of our empirical findings in Section \ref{sec:discussion}.

\section{Background}
\label{sec:back}

 \subsection{Mode-Exploring Ensembling Techniques} \label{sec:ensembling}
 In deep ensembles \citep{lakshminarayanan_simple_2017}, each model is trained on the same dataset by using different random initializations and batch orderings. As a result, the ensemble members reliably converge to different loss basins in weight-space, which makes this procedure highly effective at exploring the non-convex, multi-modal loss landscape of neural networks. Naturally, the success of deep ensembles in terms of predictive accuracy, uncertainty quantification and out-of-distribution robustness \citep{ovadia_can_2019, wenzel_hyperparameter_2020} is commonly attributed to this ability to sample from different modes that contain functionally diverse models \citep{fort_deep_2020}.
 Consequently, researchers have developed sequential ensembling methods designed to mimic the mode-sampling ability of deep ensembles. In particular, methods that leverage a cyclical (sometimes constant) learning rate schedule are supposed to explore different parts of the weight-space by cyclically increasing the learning rate and decreasing it again before sampling a model \citep{loshchilov_sgdr_2017}. Further examples are cyclical SGMCMC \citep{zhang_cyclical_2020}, as well as Snapshot Ensembles \citep{huang_snapshot_2017}, Fast Geometric Ensembling \citep{garipov_loss_2018}, and Stochastic Weight Averaging \citep{izmailov_averaging_2018}. \cite{dangelo_repulsive_2021} study ensembles with repulsive terms in weight and function space to prevent collapse to a single mode.

 \cite{frankle_linear_2020} observed that there is a point early in training after which SGD runs with different batch orderings and augmentations become \emph{linearly mode-connected}, allowing to interpolate linearly between different modes without incurring a high loss. Two models which are linearly mode-connected can be thought of as being located in the same loss basin, which, from a classical point of view, can be detrimental for the predictive diversity of an ensemble. Hence, the procedures described sequentially train models and sample from different modes by traversing regions of high loss to escape the basin and break \emph{linear mode-connectivity}.

\subsection{Permutation Hypothesis} \label{sec:permutation_hypothesis} The permutation hypothesis \citep{entezari_role_2022}, which recently has attracted much attention, posits that, for wide enough networks, the solutions obtained through independent SGD runs become linearly mode-connected if their weights are permuted appropriately. \cite{ainsworth_git_2023} provided empirical evidence for the permutation hypothesis by proposing weight and activation matching algorithms that bring neurons into alignment without altering the underlying function. Given two models found by SGD, their method finds a permutation of weights such that the two models become linearly mode connectable, i.e., lie in the same loss basin.  

The permutation hypothesis implies that the loss landscape can be collapsed into a single loss basin if we account for redundant basins that are due to permutation symmetries. This implication has significant consequences for mode-exploring ensembling techniques. The underlying assumption of such ensembling techniques is that predictive diversity can only be achieved by virtue of visiting different modes in weight-space. The permutation hypothesis, however, implies that there is nothing to be gained from visiting different modes beyond what can already be found in a single basin. In other words, visiting different basins might not be necessary to obtain models that constitute a strong ensemble.

Given these new insights into the redundancies of the loss landscape and their potential implications for deep ensembles, we depart from the dominant paradigm of exploring multiple modes and instead intentionally limit ourselves to a single mode to form a connected ensemble. By doing so, we aim to provide an answer to the following question: 

\begin{center}
    \textit{How Good Can a Single-Basin Ensemble Be?}
\end{center}

\section{Exploring a Single Basin}
\label{sec:exploring_single_basin}
\subsection{Setting}

\paragraph{Setup.} We consider image classification problems with a training dataset $\mathcal{D} = \{(\bm{x}_1, y_1), \dots, (\bm{x}_n, y_n)\}$ consisting of $n$ i.i.d tuples of labelled examples $\bm{x}_i \in \mathbb{R}^d$ and $y_i \in \{1, \dots, K\}$. We study the class of neural network functions $f_{\bm{\theta}} : \mathbb{R}^d \rightarrow \mathbb{R}^K$ parameterized by $\bm{\theta} \in \mathbb{R}^p$ where $\bm{\theta}$ denotes the concatenation of all the parameters. Note that many modern neural network parameterizations contain redundancies in the weight-space in the sense that two parameter vectors can encode the same function, i.e., $\bm{\theta}_1 \neq \bm{\theta}_2$ could yield $f_{\bm{\theta}_1}(\bm{x}) = f_{\bm{\theta}_2}(\bm{x}) \ \forall \bm{x} \in \mathbb{R}^d$. In our work, we limit ourselves to the set of symmetries induced by permutations\footnote{We note that there are other symmetries, such as scaling or sign-flip symmetries, that might cause this phenomenon. However, these other symmetries are beyond the scope of this work. }. We learn $\bm{\theta}$ through empirical risk minimization $\min_{\bm{\theta} \in \Theta} \sum_{i=1}^n \ell(f_{\bm{\theta}}(\bm{x}_i), y_i)$ with $\ell : \mathbb{R}^K \times \{1, \dots, K\} \rightarrow \mathbb{R}^{+}$ denoting the loss function. To approximately minimize this objective, we use some form of stochastic gradient descent and refer to a minimizer $\bm{\theta}$ as a \emph{mode}. Such a mode is located in a \emph{loss basin}, referring to the approximately convex region of low loss around it. We use $\text{KL}(\cdot, \cdot)$ to denote the KL-Divergence between two probability distributions over the discrete space of classification labels and denote the softmax function with $\sigma : \mathbb{R}^{K} \rightarrow (0,1)^K$.

\paragraph{Deep Ensembles.} We consider $M \in \mathbb{N}$ runs of SGD under different initializations and batch orderings, resulting in a set of minimizers $\{\bm{\theta}_1, \dots, \bm{\theta}_M\}$. Due to the non-convexity of neural landscapes, simple convex combinations $\bar{\bm{\theta}} = \sum_{i=1}^{M}\lambda_i\bm{\theta}_i$ with $\sum_{i=1}^{M}\lambda_i = 1$ do not constitute minimizers of the test loss. In the literature this is often referred to as lack of (joint) linear mode-connectivity \citep{garipov_loss_2018}. While averaging parameters proves detrimental, averaging the predictions (i.e. ensembling) leads to a substantially more powerful model, i.e. $\bar{f}(\bm{x}) := \frac{1}{M}\sum_{i=1}^{M}f_{\bm{\theta}_i}(\bm{x})$ outperforms any individual model $\bm{\theta}_i$. We demonstrate this effect on the classic image classification benchmarks CIFAR \citep{krizhevsky_learning_2009} and Tiny ImageNet \citep{le_tiny_2015} using a ResNet20 \citep{he_deep_2016} ensemble of size $M=5$ in Table \ref{tab:deep-ensemble}. While the increase in test accuracy is moderate for CIFAR10, we observe very strong improvements for the more complex tasks CIFAR100 and Tiny ImageNet. We remark that a deep ensemble of size $M$ incurs a computational cost of $C_{\text{DE}} = M \times T$ where $T$ is the number of training epochs used for a single model. We ensure that our baselines match in terms of computational cost in order to guarantee that no improvement is achieved simply due to longer training. We defer implementation details such as values for $T$ to Appendix \ref{sec:implementation_details}.

\begin{figure}
    \centering
    \includegraphics[width=0.8\textwidth]{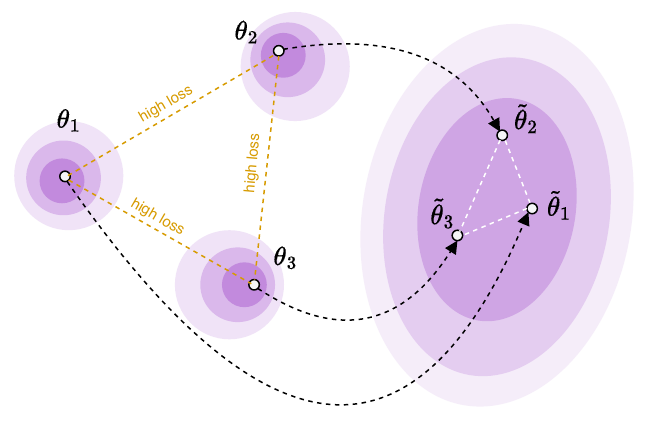}
    \caption{Illustration of toy deep ensemble $\{\bm{\theta}_1, \bm{\theta}_2, \bm{\theta}_3\}$ and the matching, connected ensemble $\{\tilde{\bm{\theta}}_1, \tilde{\bm{\theta}}_2, \tilde{\bm{\theta}}_3\}$.}
    \label{fig:connected-ensemble}
\end{figure}

\begin{table}
\centering
\begin{tabular}{lcc}
\toprule
Dataset & Mean Acc & Ens. Acc \\
\midrule
CIFAR10         & $93.01_{\pm 0.08}$ & $94.43_{\pm 0.12}$ \\[1mm]
\midrule
CIFAR100        & $73.44_{\pm 0.12}$ & $78.15_{\pm 0.10}$    \\[1mm]
\midrule
Tiny ImageNet   & $55.36_{\pm 0.33}$  & $62.85_{\pm 0.20}$  \\[1mm]

\bottomrule
\caption{Average member test accuracy $\frac{1}{5}\sum_{i=1}^{5}{\operatorname{Acc}(\bm{\theta}_i)}$ and deep ensemble performance of ResNet20. Averaged over $3$ seeds. }%
  \label{tab:deep-ensemble}
\end{tabular}
\end{table}

\subsection{Finding Connected Deep Ensembles}
\label{sec:finding-connected}
We now explore various methods to replicate the success of deep ensembles while intentionally limiting ourselves to only leveraging a single basin. In other words, we aim to construct a \textit{connected} ensemble $\{\tilde{\bm{\theta}}_1, \dots, \tilde{\bm{\theta}}_M\}$ that approximates the performance of the original ensemble while at the same time guaranteeing linear mode-connectivity. We provide a visualization of the idea in Fig.~\ref{fig:connected-ensemble}.  While for simple baselines we focus on \emph{Residual Networks} (ResNets) \citep{he_deep_2016}, we also evaluate our more competitive methods on \emph{Vision Transformers} (ViTs) \citep{dosovitskiy_image_2021}. Note that we do not consider pre-trained ViTs, as our experiments heavily depend on whether models share a pre-trained initialization. We largely focus on test accuracy, cross-entropy loss, connectivity, and diversity as the main metrics of comparison.

\paragraph{Connectivity.} In order to assess the connectivity of a given set of models $\{\bm{\theta}_1, \dots, \bm{\theta}_M\}$, we perform several tests. Firstly, we study the pairwise connectivity of any two sets of models, i.e. we evaluate 
\begin{align}
q_{\text{pair}}(\lambda) = &\operatorname{Acc}(\lambda \bm{\theta}_i + (1-\lambda)\bm{\theta}_j) \notag \\ &-(\lambda \operatorname{Acc}(\bm{\theta}_i) + (1-\lambda)\operatorname{Acc}(\bm{\theta}_j))
\end{align}
for $\lambda \in [0,1]$, where $\operatorname{Acc}:\Theta \xrightarrow{} [0,1]$ maps a configuration $\bm{\theta}$ to its generalization accuracy $\operatorname{Acc}(\bm{\theta})$. If $q_{\text{pair}}$ does not decrease significantly, we say that $\{\bm{\theta}_1, \dots, \bm{\theta}_M\}$ are pairwise linearly-connected. We further visualize the test accuracy as a function of $\lambda$ (see e.g. Fig.~\ref{fig:lmc}) and inspired by \citet{garipov_loss_2018}, also produce two-dimensional plots visualizing the parameter-plane spanned by three models (see e.g. Fig.~\ref{fig:weight_space_cuts_tiny_imagenet}). In order to assess joint connectivity, we measure 
\begin{align}q_{\text{joint}}(\bm{\lambda}) = \operatorname{Acc}\left(\sum_{i=1}^{M} \lambda_i \bm{\theta}_i\right)  -\sum_{i=1}^{M}\lambda_i \operatorname{Acc}(\bm{\theta}_i), \hspace{1mm} \bm{\lambda} \sim \operatorname{Dir}(\bm{1})
\end{align}
While we could rely on a densely sampled grid for $\lambda$ in the pairwise case, for the joint connectivity we resort to randomly sampling convex combination weights $\bm{\lambda} \in [0,1]^{M}$ from a Dirichlet distribution. We then average multiple draws of $\bm{\lambda}$ to obtain $\bar{q}_{\text{joint}}$ and say that $\{\bm{\theta}_1, \dots, \bm{\theta}_M\}$ are jointly linearly-connected if $\bar{q}_{\text{joint}}$ does not decrease significantly.

 \paragraph{Diversity.} To evaluate ensemble diversity, we consider two metrics used by \cite{abe_pathologies_2023}. More specifically, we consider the variance over ensemble members' true-class predictions $\operatorname{Var}(\{\bm{\theta}_i\}_{i=1}^M)$ and the one-vs-all Jensen-Shannon-Divergence $\operatorname{JSD}(\{\bm{\theta}_i\}_{i=1}^M)$. For more details on those metrics, we refer to \cite{abe_pathologies_2023}.

\paragraph{Stochastic Weight Ensembling (SWE).} As a very simple first baseline, we consider a variant of \textit{stochastic weight averaging} (SWA) \citep{izmailov_averaging_2018}, where instead of averaging the obtained iterates, we average the predictions, effectively forming an ensemble. Similarly to \citet{izmailov_averaging_2018}, we use a decaying learning rate schedule that converges to a fixed value, enabling exploration of the basin without leaving it. More precisely, we train a \textit{ResNet20} for $T$ epochs with a decaying learning rate, producing the first sample $\tilde{\bm{\theta}}_1$ and then continue training with a constant learning rate, saving a checkpoint $\tilde{\bm{\theta}}_i$ every $T$ epochs until we collected $M$ samples. This ensures the same computational budget as the deep ensemble. We highlight that this approach is very similar to Snapshot Ensembles \citep{huang_snapshot_2017}, but instead of encouraging explorations of different basins by using cyclical learning rates, we ensure connectivity by using a constant learning rate. We report the resulting test performance and joint connectivity values in Table \ref{tab:constrained_ensembles}. We also display the corresponding values for deep ensembles as a reference. We observe that SWE is surprisingly effective, matching the performance of the deep ensemble on CIFAR10 while maintaining a high degree of connectivity. On the more challenging task of CIFAR100 however, we find a significant gap of roughly $3\%$ in test accuracy.

\paragraph{Constrained Ensembles.} We leverage the insights of \citet{frankle_linear_2020} regarding the stability of SGD; Along the training trajectory  $\{\bm{\theta}^{(t)}: t\leq T\}$ of SGD, there exists a point $\bm{\theta}^{(t)}$ after which any subsequently started SGD run with a different batch ordering ends up in the same loss basin. Surprisingly, this time point $t$ can be as early as a few epochs in training. This offers a recipe for the following family of connected ensembles; (1) Train a model up to time $t$. (2) Use $\bm{\theta}^{(t)}$ as a starting point for $M$ runs of SGD for $T-t$ epochs with different batch orderings, leading to a \textit{constrained} ensemble $\{\tilde{\bm{\theta}}_1(t), \dots, \tilde{\bm{\theta}}_M(t)\}$. Again we use the same computational budget as a deep ensemble.  We notice that the time parameter $t$ intuitively trades off diversity and connectivity; the smaller $t$ is, the more diverse but less connected are the solutions and similarly for large $t$. We display the resulting performance and connectivity results in Table \ref{tab:constrained_ensembles}. We obtain a very similar picture as for SWE, i.e., constrained ensembles also match the performance on CIFAR10, offer strong connectivity, but fall short on CIFAR100, albeit with a significant improvement.
\input{tables/constrained_ensembles}

\section{Re-discovering Deep Ensembles in a Single Basin}
\label{sec:rediscovering_single_basin}
Our preliminary results lead us to conclude that discovering a connected deep ensemble with matching performance is a challenging endeavour. We thus take a step back and revisit our research question from a slightly different angle: 
\begin{center}
    \textit{If access to a deep ensemble was granted, could one re-discover it in a single basin?}
\end{center}

This is conceptually a simplified goal as knowledge of other basins can now be leveraged to guide the search within a single basin. A positive answer however would still be very impactful as it proves the existence of a connected deep ensemble, motivating further research into efficient exploration of a single basin. \\[2mm] In the following approaches, we will thus assume that we have access to a deep ensemble $\{\bm{\theta}_1, \dots, \bm{\theta}_{M}\}$. We emphasize that the purpose of this investigation is \emph{not} to propose an alternative method for ensembling. Instead, we aim to investigate how constraining ensembles to a single loss basin impacts predictive performance and calibration to gain a better understanding of the loss landscapes of neural networks.

\subsection{Permuting Deep Ensembles}

\paragraph{Pairwise Alignment.} As a first candidate, we investigate the \textsc{PermutationCoordinateDescent} (PCD) algorithm from \cite{ainsworth_git_2023}. We choose the first member $\bm{\theta}_1$ as a reference model and we aim to apply permutations $\bm{\pi}_i$ to each remaining member $\bm{\theta}_i$ such that $\bm{\theta}_1$ and $\bm{\pi}_i(\bm{\theta}_i)$ live in the same basin. Such a permutation is discovered by aligning the weights of each member with the reference model, we refer to \citet{ainsworth_git_2023} for more details. Since permutations constitute a symmetry of neural networks, the performance of the new members $\bm{\pi}_i(\bm{\theta}_i)$ remains unchanged, and we thus have a mathematical guarantee to achieve the same performance as the original ensemble. Unfortunately, this guarantee comes at a cost; while $\bm{\theta}_1$ and $\bm{\pi}(\bm{\theta}_i)$ are indeed linearly-connected for $i=1, \dots, M$ as demonstrated in \citet{ainsworth_git_2023}, pairwise connectivity between two permuted members $\bm{\pi}_i(\bm{\theta}_i)$ and $\bm{\pi}_j(\bm{\theta}_j)$ does not hold. Similarly, joint connectivity is also violated as shown in Table \ref{tab:pcd_multi_pcd}. This is not surprising as the objective only optimizes for pairwise alignment.

\begin{table}
\centering
\resizebox{1\linewidth}{!}{
\begin{tabular}{@{}lrcrcr@{}}\toprule

& Deep Ens.  & PCD  & Multi-PCD\\

\midrule

CIFAR10                    & $-71.74_{\pm 2.38}$ & $-25.84_{\pm 4.20}$ & $-14.64_{\pm 3.66}$\\
CIFAR100                   & $-68.16_{\pm 1.72}$ & $-44.89_{\pm 0.91}$ & $-41.02_{\pm 1.55}$\\
Tiny ImageNet              & $-53.78_{\pm 0.85}$ & $-46.30_{\pm 2.08}$ & $-44.54_{\pm 2.65}$\\
\bottomrule
\end{tabular}
}
\caption{Joint connectivity $\bar{q}_{\text{joint}}$ of deep ensembles and permuted ensembles optimizing for pairwise (PCD) and joint alignment (Multi-PCD). Averaged over $3$ seeds.}
\label{tab:pcd_multi_pcd}
\end{table}

\paragraph{Joint Alignment.} Next, we evaluate whether the lack of joint connectivity can be diminished by extending the optimization objective used in PCD. More specifically, we change the objective function used in \cite{ainsworth_git_2023} to account for the joint alignment with respect to all other models and not just the reference model.  Thus, when optimizing $\bm{\pi}_i(\bm{\theta}_i)$ we account for the alignment with respect to all other models $\bm{\pi}_j(\bm{\theta}_j)$ with $j \neq i$ in the ensemble. Using this modified objective and wrapping the pairwise procedure with another layer iterating over ensemble members, we obtain an algorithm that optimizes for joint alignment and to which we refer to as Multi-PCD.  While joint connectivity does improve, the resulting ensemble is still far from being connected as measured by $\bar{q}_{\text{joint}}$ in Table~\ref{tab:pcd_multi_pcd}. We thus conclude that permutations can not be leveraged to re-discover an ordinary multi-basin ensemble in a single loss basin.

\subsection{Distilling Deep Ensembles}

\begin{figure}[h]
     \centering
     \includegraphics[width=\textwidth]{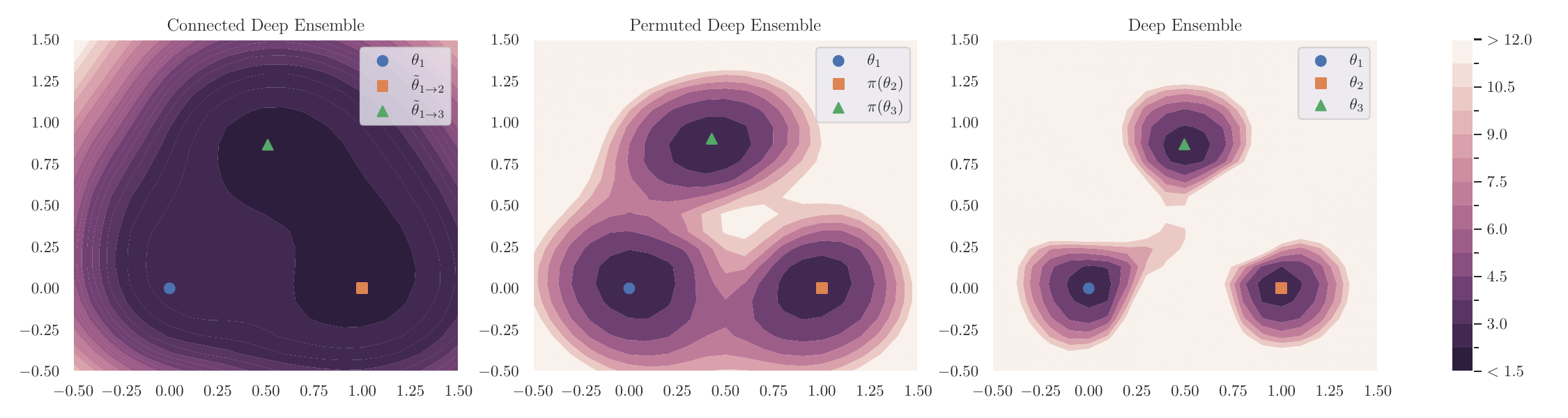}
    \caption{The plots display the 2D planes spanned by the three weight vectors given by the parameters of a ResNet20 trained on Tiny ImageNet mentioned in the legend with the first model at the origin. The plane is constructed as in \cite{garipov_loss_2018}.}
    \label{fig:weight_space_cuts_tiny_imagenet}
\end{figure}

\paragraph{Distilled Ensemble.} 
In this approach, we combine our insights from \textit{constrained} ensembles with the mechanism of model distillation, as introduced by \citet{hinton_distilling_2015}. Again denote by $\bm{\theta}_1^{(t)}$ the stability point of SGD for the reference model $\bm{\theta}_1$. We aim to re-discover the $j$-th member $\bm{\theta}_j$ in the same basin as $\bm{\theta}_1$ by minimizing a distillation objective towards $\bm{\theta}_j$, i.e. we minimize 
\begin{align}
    \mathcal{L}(\bm{\theta}) 
    =& \sum_{i=1}^n (1-\beta) \cdot \tau^2 \cdot \text{KL}\left(\sigma\left(\frac{f_{\bm{\theta}_j}(\bm{x}_i)}{\tau}\right),\sigma\left(\frac{f_{\bm{\theta}}(\bm{x}_i)}{\tau}\right)\right) \notag\\ &-\beta  \log \left(\left[\sigma\left(f_{\bm{\theta}}(\bm{x}_i)\right)\right]_{y_i}\right) 
    \label{eq:distill-obj}
\end{align}
\begin{figure}[h]
      \begin{subfigure}[b]{0.49\textwidth}
         \centering
         \includegraphics[width=\textwidth]{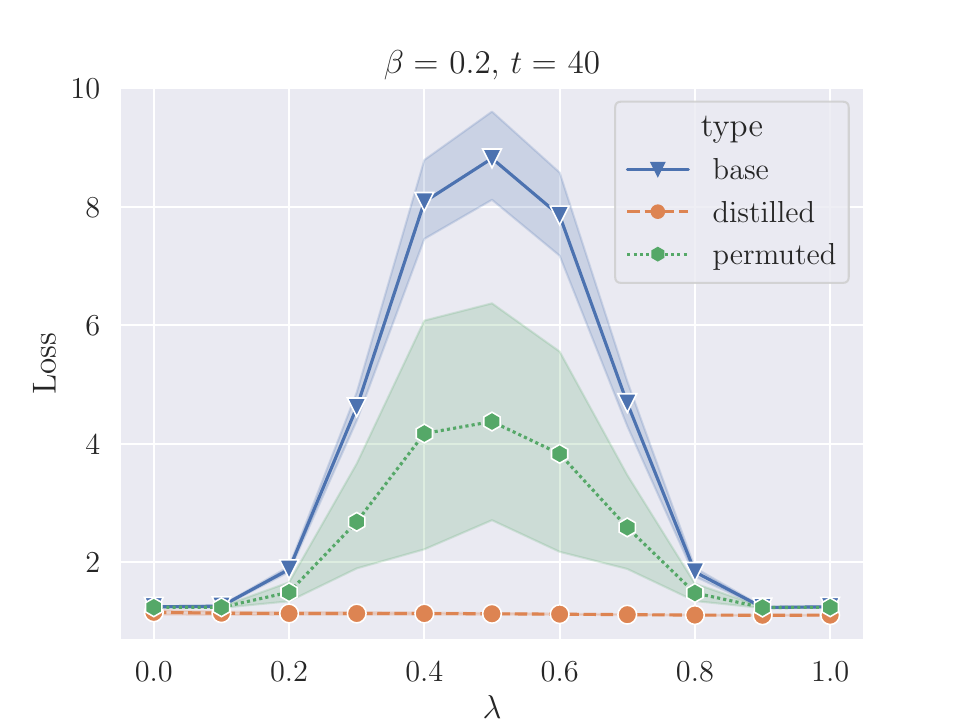}
         \caption{ResNet20, CIFAR100}
         \label{fig:lmc_resnet_cifar100}
     \end{subfigure}
     \hfill
     \begin{subfigure}[b]{0.49\textwidth}
         \centering
         \includegraphics[width=\textwidth]{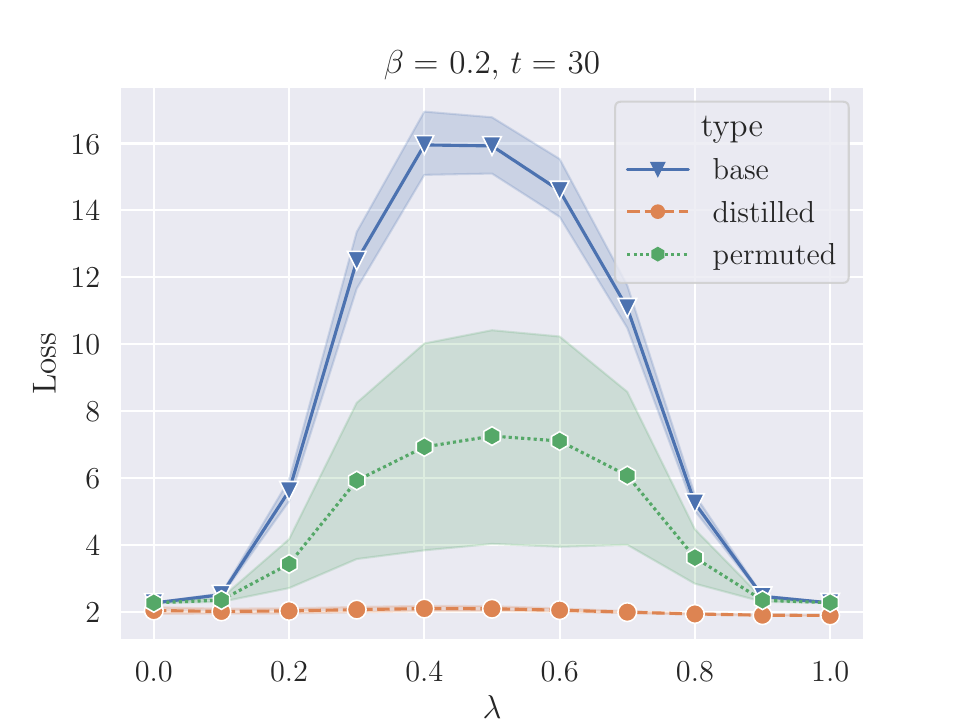}
         \caption{ResNet20, Tiny ImageNet}
         \label{fig:lmc_resnet_tiny}
     \end{subfigure}

      \begin{subfigure}[b]{0.49\textwidth}
         \centering
         \includegraphics[width=\textwidth]{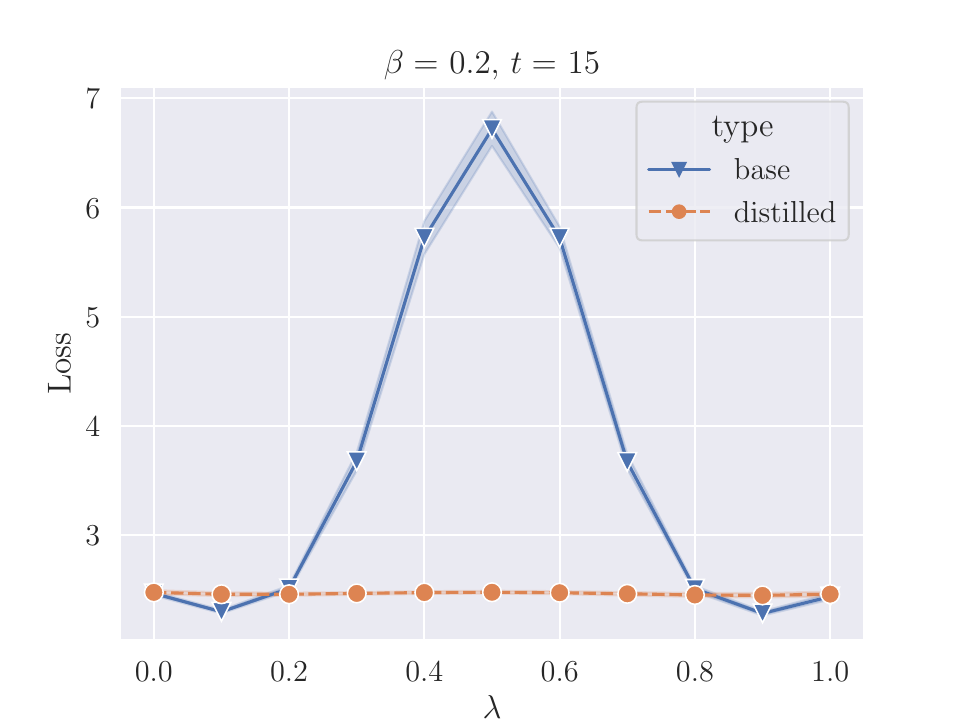}
         \caption{ViT, CIFAR100}
         \label{fig:lmc_vit_cifar100}
     \end{subfigure}
     \hfill
     \begin{subfigure}[b]{0.49\textwidth}
         \centering
         \includegraphics[width=\textwidth]{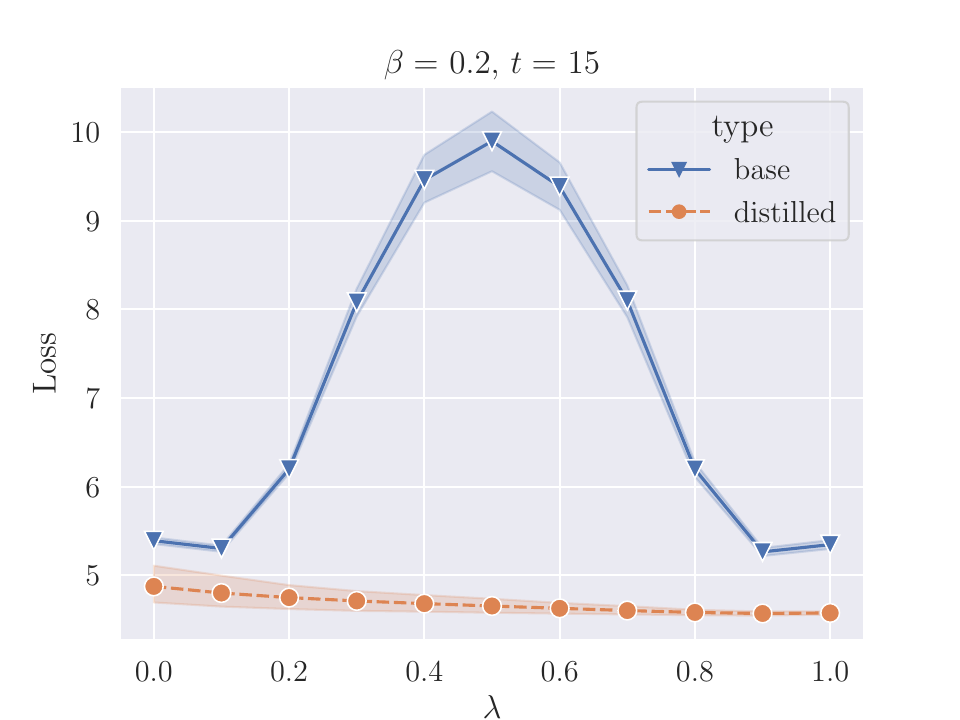}
         \caption{ViT, Tiny ImageNet}
         \label{fig:lmc_vit_tiny}
     \end{subfigure}
    \caption{Linear Mode Connectivity of ResNet20 and ViT ensembles. We approximate $q_{pair}$ through lines showing averages of five randomly selected pairs. The experiment is repeated with three random seeds, totalling 15 pairs. The shading shows the standard deviation.}
    \label{fig:lmc}
\end{figure}

where $\beta$ and $\tau$ are additional hyperparameters. We then start the optimization from $\bm{\theta}_1^{(t)}$ to encourage connectivity of solutions and denote the minimizers of  Eq.~\ref{eq:distill-obj} by $\tilde{\bm{\theta}}_{1 \text{ \Arrow{0.15cm}} j}$. $\beta$ trades-off the optimization towards matching the ground truth and functional similarity to the $j$-th member. Note that for $\beta=1$, the approach essentially reduces to \textit{constrained} ensembles. $\tau > 1$ is the temperature parameter commonly used in knowledge distillation frameworks to increase the entropy of the distribution over class labels, facilitating the knowledge transfer to the student model. Table~\ref{tab:beta_table} illustrates that our distillation strategy with $\beta=0.2$ produces very competitive ensembles for residual models, significantly closing the gap to standard deep ensembles across all datasets. Moreover, such a distilled ensemble exhibits a surprisingly strong degree of connectivity, not only fulfilling pairwise connectivity (see Fig.~\ref{fig:lmc}), but also the very challenging joint linear connectivity $\bar{q}_{\text{joint}}$, as shown in Fig.~\ref{fig:weight_space_cuts_tiny_imagenet} and Table \ref{tab:beta_table}.  

\begin{figure}
    \centering
    \includegraphics[width=\textwidth]{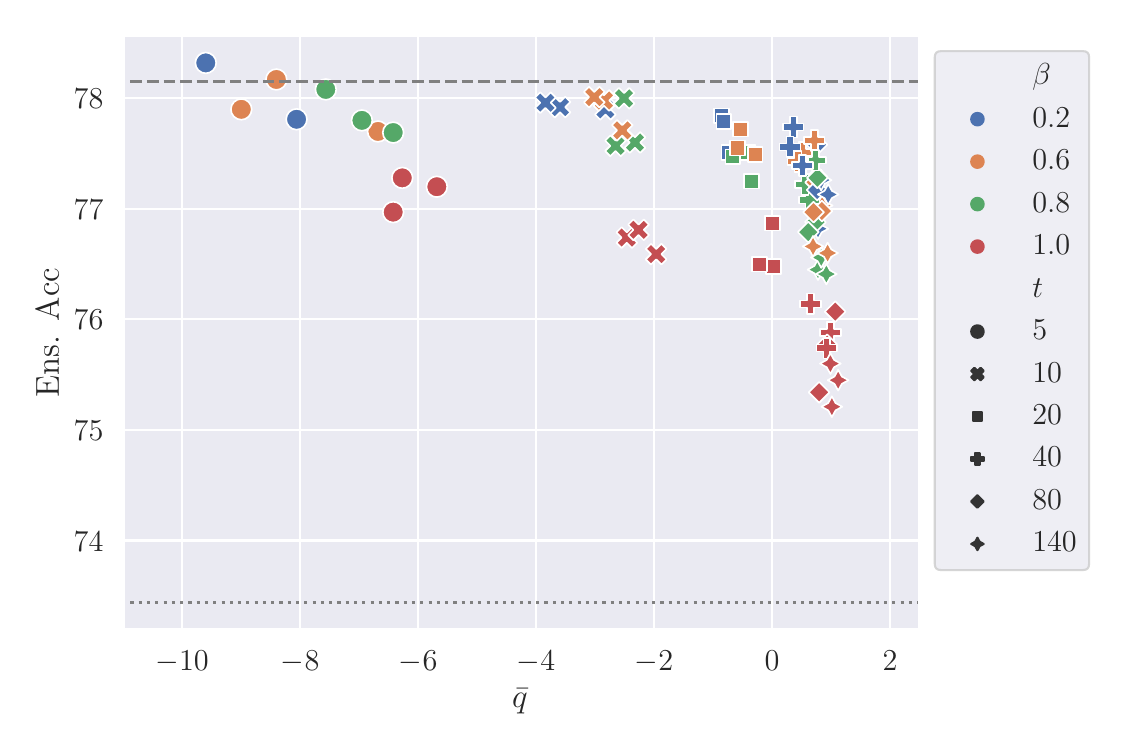}
    \caption{Connectivity $\bar{q}$ plotted against test accuracy for ResNet20 on CIFAR100. The dashed horizontal line shows the accuracy of a deep ensemble, while the dotted horizontal line shows the mean member accuracy.}
    \label{fig:con-acc}
\end{figure}

Note that we do not explicitly encourage similarity between two distilled points $\tilde{\bm{\theta}}_{1 \text{ \Arrow{0.15cm}} i}$ and $\tilde{\bm{\theta}}_{1 \text{ \Arrow{0.15cm}} j}$. However, the stability of SGD after time $t$ suffices to guarantee joint connectivity. In contrast to the weight-matching procedure introduced above, the distillation framework can also readily be applied to any architecture, including architectures with non-elementwise non-linearities such as Vision Transformers.

\input{tables/beta1.0_vs_beta0.2}

\begin{figure}
    \centering
    \begin{subfigure}[b]{0.49\textwidth}
        \includegraphics[width=\linewidth]{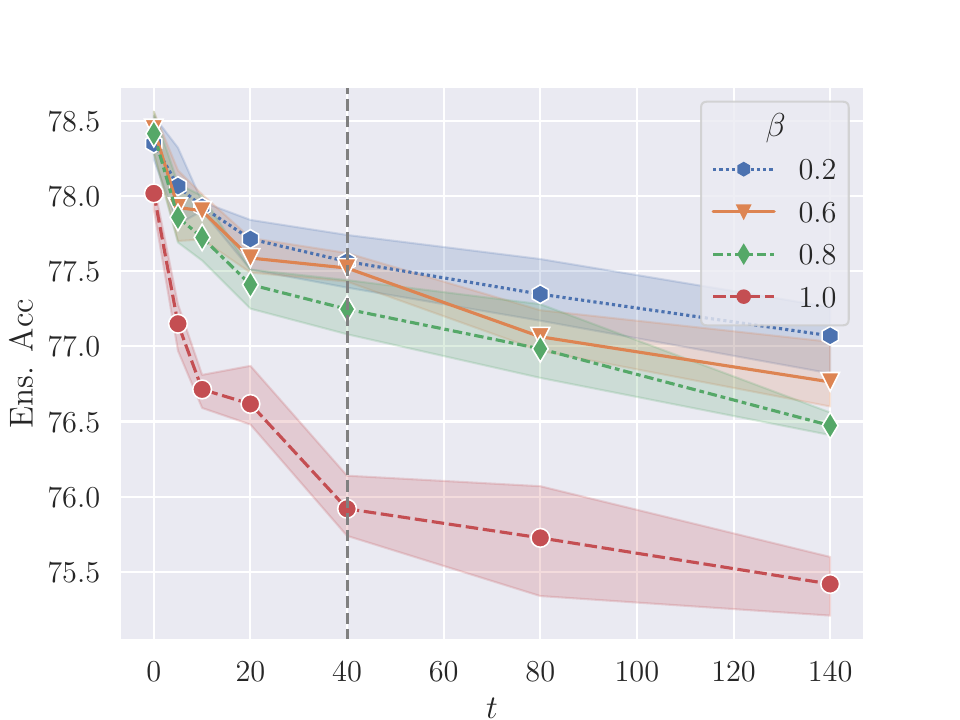}
        \caption{Ensemble Accuracy}
        \label{fig:t-ablation-accuracy}
    \end{subfigure}
    \hfill
    \centering
    \begin{subfigure}[b]{0.49\textwidth}
        \includegraphics[width=\linewidth]{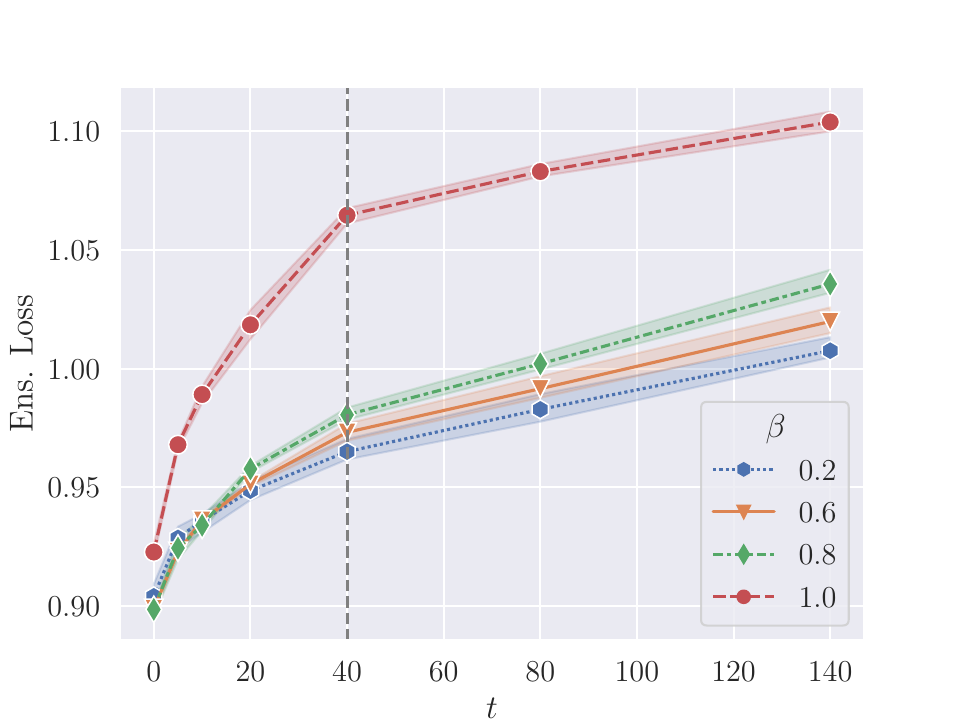}
        \caption{Ensemble Loss}
        \label{fig:t-ablation-loss}
    \end{subfigure}
    \caption{Accuracy, loss, and mean accuracy as a function of time parameter $t$ for ResNet20 on CIFAR100. The dashed vertical lines mark the $t$ used in Table~\ref{tab:beta_table}.}
    \label{fig:t-ablation}
\end{figure}

\paragraph{Role of $\beta$ and $t$.} We now investigate the impact of the distillation procedure by varying the hyperparameter $\beta$ in Eq. \ref{eq:distill-obj}. When $\beta=1$, our training procedure reduces to standard training restricted to a single basin, recovering the constrained ensembles from Section~\ref{sec:finding-connected}. In contrast, when $0 < \beta < 1$, we optimize the convex combination of the cross-entropy loss and a similarity encouragement term relative to an out-of-basin model. The results in Table \ref{tab:beta_table} demonstrate that similarity encouragement to an out-of-basin model significantly boosts the accuracy of the connected ensemble for both architectures considered. We further highlight the important role of the splitting time $t$ in Fig.~\ref{fig:t-ablation}, trading-off connectivity and performance. Also, note that, as expected, the deterioration in performance as connectivity increases is more pronounced for ensembles that do not include a distillation term ($\beta=1.0$).

\paragraph{Connectivity and Accuracy.} In Fig.~\ref{fig:con-acc}, we show test accuracy as a function of connectivity $\bar{q}$. Without distillation (represented by the red markers), we see that increased connectivity negatively impacts performance. However, as soon as we employ distillation (e.g., blue markers), we manage to significantly mitigate the drop in performance without compromising connectivity, narrowing the gap to the baseline of deep ensembles.

\paragraph{Isolating the regularizing effect of distillation.} \cite{zhang_be_2019, furlanello_born_2018} have demonstrated that distillation can enhance student performance, to the point that the student can even surpass the performance of its teacher. Thus, the improvement of distilled ensembles beyond the performance of constrained ensembles observed in Table~\ref{tab:beta_table} might be caused by this regularizing effect of distillation. To clearly isolate the magnitude of this effect, we consider a baseline of a deep ensemble trained with the same distillation objective in Eq.~\ref{eq:distill-obj}. If the gain in ensemble performance observed for distilled ensembles would be primarily due to this regularizing effect of distillation rather than the incorporation of out-of-basin knowledge, then it is reasonable to expect similar improvements in ensemble performance when adding the distillation term to deep ensembles. As illustrated in Table~\ref{tab:isolating_distillation}, distillation does not significantly improve ensemble performance for ordinary deep ensembles, highlighting that the gain observed in Table~\ref{tab:beta_table} is unlikely to be caused by the regularizing effect of distillation.

\input{tables/isolating_distillation}

\begin{figure}
    \centering
    \begin{subfigure}[b]{0.49\textwidth}
        \includegraphics[width=\textwidth]{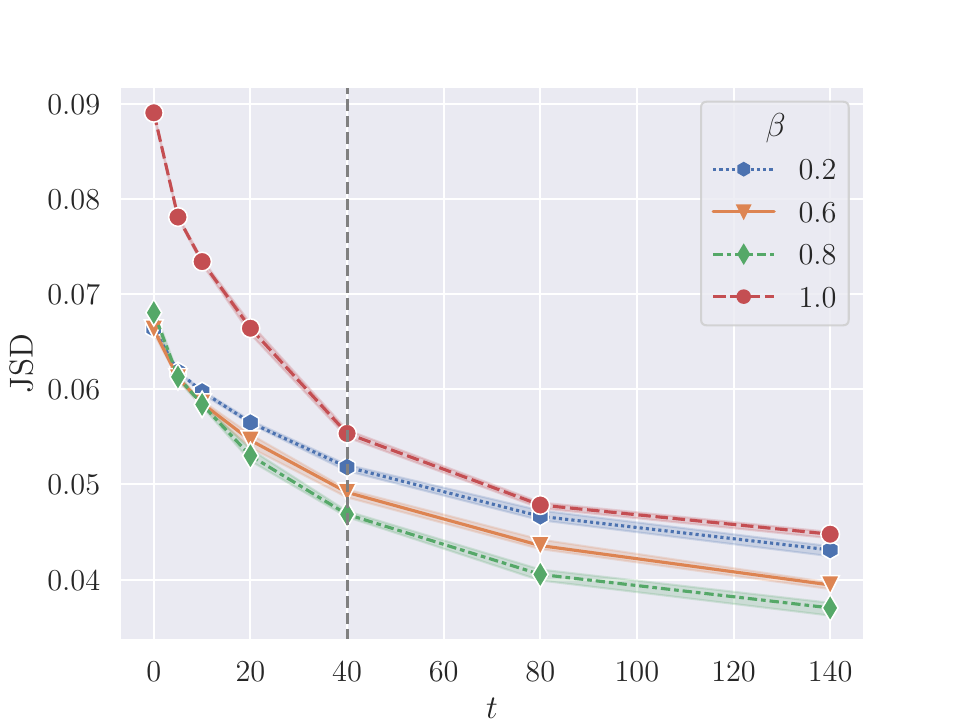}
        \caption{JSD}
        \label{fig:t-ablation-jsd}
    \end{subfigure}
    \centering
    \begin{subfigure}[b]{0.49\textwidth}
        \includegraphics[width=\textwidth]{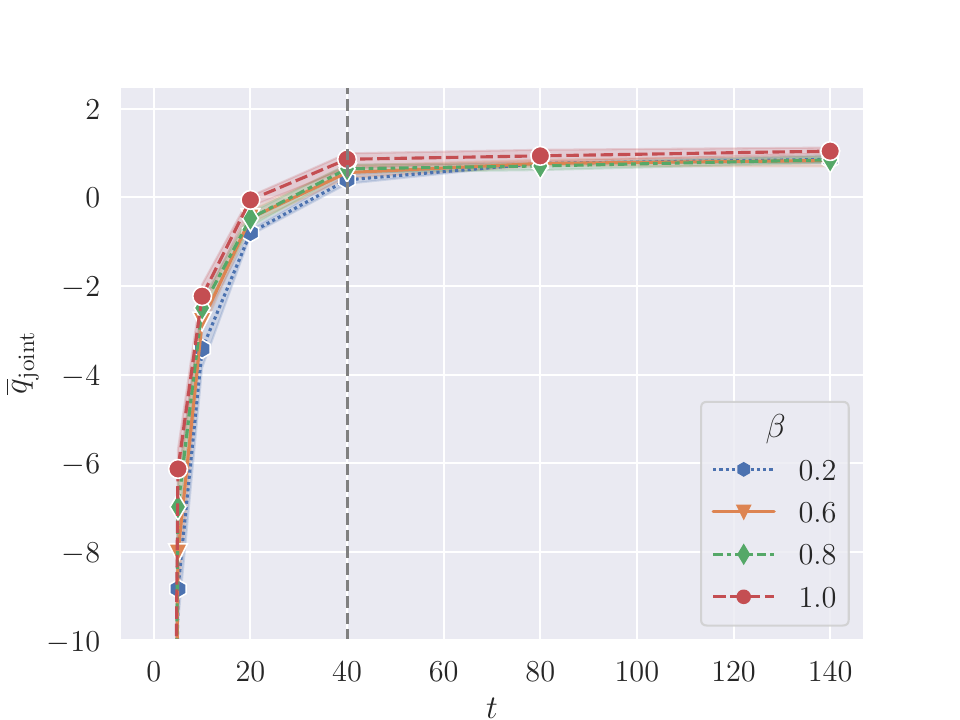}
        \caption{$\bar{q}_{\text{joint}}$}
        \label{fig:t-ablation-connectivity}
    \end{subfigure}
    \caption{Jensen-Shannon divergence and joint connectivity as a function of time parameter $t$ for ResNet20 ensembles on CIFAR100. The dashed vertical lines mark the $t$ used in Table~\ref{tab:beta_table}.}
    \label{fig:t-ablation-div-con}
\end{figure}

\paragraph{Diversity and Connectivity.} In search of a suitable explanation for the performance gap between connected and deep ensembles, we turn to predictive diversity, which is often deemed as the primary driver of ensemble performance \citep{dietterich_ensemble_2000, breiman_bagging_1996, freund_short_1999}.  In Fig.~\ref{fig:t-ablation-div-con}, we plot diversity and connectivity as a function of $t$ for a grid of $\beta$ values. In the limit of $t \rightarrow 0$, we recover ordinary deep ensembles, due to the loss of joint connectivity. In contrast, for $t > 40$, we enter the regime of connected ensembles whose members fulfill joint connectivity. Note that with increasing connectivity, the diversity, as measured by the one-vs-all Jensen-Shannon Divergence decreases, giving rise to a \textit{diversity-connectivity trade-off} that is likely to be a main driver of the performance gap between single-basin and ordinary multi-basin ensembles.

\section{Related Works} 
\label{sec:related-work}

\paragraph{Connected Ensembles.} There is a plethora of previous work that studies novel ensembling techniques, often with a focus on reduced cost or better weight averaging properties. \citet{jolicoeur-martineau_population_2023} propose an algorithm that simultaneously trains all ensemble members, dynamically updating the members with a running average. \citet{garipov_loss_2018} and \citet{huang_snapshot_2017} both adapt a similar strategy as the SWE approach but use a cyclical learning rate to intentionally break connectivity and produce more efficient ensembles. \citet{wortsman_learning_2021} on the other hand directly learn lines and curves whose endpoints they leverage for ensembling. They also report improved performance when using the midpoint as a summary of the ensemble. Another related line of work studies fusion of several independent models. \citet{singh_model_2020} leverage optimal transport to align the weights of multiple models and produce a fused endpoint. \citet{ainsworth_git_2023} take a similar approach and fuse different networks by finding fitting permutations to maximize similarity. We provide a more detailed overview of related techniques in Appendix~\ref{sec:more-related_work} due to space constraints.

\paragraph{Mode Connectivity.} An intellectual ancestor to linear mode connectivity can be seen in the work of \citep{goodfellow_qualitatively_2015}. They consider the 1D subspace spanned by the initial and fully trained parameter vectors and find that the loss is monotonically decreasing the closer we get to the final parameter vector. \citep{lucas_monotonic_2021} confirmed these results and coined the phenomenon \textit{monotonic linear interpolation}. In the context of our work, we interpret this monotonic linear interpolation phenomenon as a descent into a loss basin whose functional diversity we aim to explore. \cite{frankle_linear_2020} demonstrated that there is a point in training $\bm{\theta}^{(t)}$ after which SGD runs sharing $\bm{\theta}^{(t)}$ as initialization remain linearly mode connected. \cite{neyshabur_what_2020} observed linear mode connectivity in a transfer learning setup, where models pre-trained on a source task remain linearly mode connected after training on the downstream task. \cite{juneja_linear_2023} provide counterexamples to mode connectivity outside of image classification tasks. \cite{draxler_essentially_2018, garipov_loss_2018} found non-linear paths of low loss between independently trained models, questioning the idea that the loss landscape is composed of isolated minima.

\section{Discussion}
\label{sec:discussion}
In this work, we have explored various approaches to construct ensembles constrained to lie in a single basin. We observe that constructing such a connected ensemble without any knowledge from other basins proves to be difficult and a significant gap to deep ensembles remains. Moreover, we observe a pronounced trade-off between (joint) linear mode-connectivity and the resulting ensemble performance and diversity.  However, when incorporating other basins implicitly through a distillation procedure we manage to break this trade-off and strongly reduce this gap, producing connected ensembles that are (almost) on-par for convolutional networks. While relying on other basins renders our approach very inefficient, it nevertheless proves the existence of very performant ensembles in a single basin, requiring us to rethink the characteristics of loss landscapes. We remark however that the picture is less clear for architectures with less inductive bias such as ViTs.

The existence of strong connected ensembles demonstrates that, in principle, the functional diversity within a single basin is sufficient to achieve predictive performance that is comparable to an ensemble sampled from different modes. In other words, our results illustrate that escaping the basin is not a prerequisite for attaining competitive prediction accuracy. We hope that our insights can guide future work towards designing algorithms that more thoroughly and efficiently explore a single basin.

\bibliographystyle{apalike}
\bibliography{references}

\appendix
\clearpage

\input{appendix}

\end{document}

%% file: extrapackages.tex
\usepackage{tikz}
\usepackage{hyperref}
\usepackage{url}

\usepackage{appendix}
\usepackage[utf8]{inputenc} 
\usepackage[T1]{fontenc}    
\usepackage{hyperref}       
\hypersetup{
    colorlinks=True,
    linkcolor=red,
    filecolor=green,      
    urlcolor=blue,
    citecolor=blue
}
\usepackage{wrapfig,lipsum,booktabs}
\usepackage{floatrow}
\usepackage{url}            
\usepackage{booktabs}       
\usepackage{amsfonts}       
\usepackage{nicefrac}       
\usepackage{microtype}      
\usepackage{xcolor}         

\usepackage{booktabs}
\usepackage{multirow}
\usepackage{bm}
\usepackage{mathtools}
\usepackage{caption}
\usepackage{subcaption}
\usepackage{wrapfig}

\usepackage{algorithm}
\usepackage{algpseudocode}

\usepackage{multibib}

\newfloatcommand{capbtabbox}{table}[][\FBwidth]

\newcommand{\Arrow}[1]{%
\parbox{#1}{\tikz{\draw[->](0,0)--(#1,0);}}
}

%% file: tables/constrained_ensembles.tex
\begin{table*}[!h]
\centering
\resizebox{1\linewidth}{!}{
\begin{tabular}{@{}lrrcrrcrr@{}}\toprule
& \multicolumn{2}{c}{Deep Ensemble} & \phantom{abc} & \multicolumn{2}{c}{Constrained Ensemble} & \phantom{abc}& \multicolumn{2}{c}{SWE}\\
\cmidrule{2-3} \cmidrule{5-6} \cmidrule{ 8-9}

& $\bar{q}_{\text{joint}}$ & Ens. Acc  && $\bar{q}_{\text{joint}}$ & Ens. Acc && $\bar{q}_{\text{joint}}$ & Ens. Acc\\

\midrule

 C10        & $-71.74_{\pm 2.38}$ & $94.43_{\pm 0.12}$ && $-0.14_{\pm 0.07}$ & $94.17_{\pm 0.05}$ && $1.48_{\pm 0.04}$ & $94.00_{\pm 0.18}$\\
 C100       & $-68.16_{\pm 1.72}$ & $78.15_{\pm 0.10}$ && $0.86_{\pm 0.18}$ & $75.92_{\pm 0.20}$  && $3.30_{\pm 0.10}$ & $74.95_{\pm 0.49}$\\
Tiny-IN   & $-53.78_{\pm 0.85}$ & $62.85_{\pm 0.20}$ && $0.75_{\pm 0.10}$ & $59.83_{\pm 0.13}$  && $2.80_{\pm 0.14}$ & $58.36_{\pm 0.60}$\\

\bottomrule
\end{tabular}
}
\caption{Comparison of ResNet20 ensemble connectivity and accuracy (in percent) for deep ensembles and connected ensembles.  Averaged over $3$ seeds.}
\label{tab:constrained_ensembles}
\end{table*}

%% file: tables/beta1.0_vs_beta0.2.tex
\begin{table*}[!ht]
\centering

\resizebox{1\linewidth}{!}{
\begin{tabular}{@{}llrrrcrrr@{}}\toprule
&& \multicolumn{3}{c}{$\beta=1.0$} & \phantom{ab}  & \multicolumn{3}{c}{$\beta=0.2$}\\
\cmidrule{3-5} \cmidrule{7-9} 

&& $\bar{q}_{\text{joint}}$ & Mean Acc & Ens. Acc  && $\bar{q}_{\text{joint}}$ & Mean Acc & Ens. Acc\\

\midrule
\multirow{2}{*}{CIFAR10}

& ResNet20  & $-0.14_{\pm 0.07}$ & $93.15_{\pm 0.03}$ & $94.17_{\pm 0.05}$ && $-0.64_{\pm 0.11}$ & $93.67_{\pm 0.12}$ &  $94.46_{\pm 0.20}$\\
& ViT       & $-1.37_{\pm 0.41}$ & $82.60_{\pm 0.02}$& $84.28_{\pm 0.23}$  && $-1.49_{\pm 0.25}$ & $83.14_{\pm 0.13}$ & $84.55_{\pm 0.40}$\\

\midrule
\multirow{2}{*}{CIFAR100}

& ResNet20  & $0.86_{\pm 0.18}$ & $73.53_{\pm 0.23}$ & $75.92_{\pm 0.20}$   && $0.39_{\pm 0.11}$& $75.33_{\pm 0.12}$  & $77.56_{\pm 0.18}$\\
& ViT       & $-0.14_{\pm 0.08}$ & $54.90_{\pm 0.26}$ & $57.81_{\pm 0.29}$   && $-0.29_{\pm 0.33}$& $56.12_{\pm 0.10}$  & $58.70_{\pm 0.15}$\\

\midrule
\multirow{2}{*}{Tiny ImageNet}

& ResNet20  & $0.75_{\pm 0.10}$ & $55.80_{\pm 0.19}$ & $59.83_{\pm 0.13}$  && $-1.35_{\pm 0.48}$ & $58.69_{\pm 0.17}$ & $62.61_{\pm 0.43}$\\
& ViT       & $1.76_{\pm 0.12}$ & $35.36_{\pm 0.30}$ & $39.50_{\pm 0.21}$  && $1.57_{\pm 0.18}$  & $38.46_{\pm 0.07}$  & $42.31_{\pm 0.09}$\\

\bottomrule

\end{tabular}
}
\caption{Comparison of joint connectivity and ensemble performance for constrained ($\beta=1.0$) and distilled ensembles ($\beta=0.2$).  Averaged over $3$ seeds.}
\label{tab:beta_table}
\end{table*}

%% file: tables/isolating_distillation.tex
\begin{table*}[!ht]
\centering

\resizebox{1\linewidth}{!}{
\begin{tabular}{@{}llrrrcrrr@{}}\toprule
&& \multicolumn{3}{c}{Deep Ens.} & \phantom{ab}  & \multicolumn{3}{c}{Deep Ens. + $\beta=0.2$}\\
\cmidrule{3-5} \cmidrule{7-9} 

&& $\bar{q}_{\text{joint}}$ & Mean Acc & Ens. Acc  && $\bar{q}_{\text{joint}}$ & Mean Acc & Ens. Acc\\

\midrule
\multirow{2}{*}{CIFAR10}

& ResNet20  & $-71.74_{\pm 2.38}$ & $93.01_{\pm 0.08}$ & $94.43_{\pm 0.12}$ && $-71.30_{\pm 3.01}$ & $93.54_{\pm 0.04}$ &  $94.45_{\pm 0.02}$\\
& ViT       & $-55.81_{\pm 1.99}$ & $82.43_{\pm 0.33}$& $85.10_{\pm 0.27}$  && $-55.70_{\pm 1.71}$ & $82.97_{\pm 0.22}$ & $84.87_{\pm 0.31}$\\

\midrule
\multirow{2}{*}{CIFAR100}

& ResNet20  & $-68.16_{\pm 1.72}$ & $73.44_{\pm 0.12}$ & $78.15_{\pm 0.10}$   && $-69.03_{\pm 2.19}$& $75.20_{\pm 0.15}$  & $78.42_{\pm 0.20}$\\
& ViT       & $-47.28_{\pm 0.19}$ & $54.91_{\pm 0.10}$ & $59.88_{\pm 0.12}$   && $-48.32_{\pm 0.15}$& $56.20_{\pm 0.08}$  & $59.92_{\pm 0.26}$\\

\midrule
\multirow{2}{*}{Tiny ImageNet}

& ResNet20  & $-53.78_{\pm 0.85}$ & $55.36_{\pm 0.33}$ & $62.85_{\pm 0.20}$  && $-56.54_{\pm 0.70}$ & $58.65_{\pm 0.23}$ & $63.29_{\pm 0.33}$\\
& ViT       & $-33.04_{\pm 0.70}$ & $35.57_{\pm 0.38}$ & $44.05_{\pm 0.19}$  && $-35.79_{\pm 0.77}$  & $38.37_{\pm 0.31}$  & $44.29_{\pm 0.21}$\\

\bottomrule

\end{tabular}
}
\caption{Isolating the additional regularizing effect of distillation.  Averaged over $3$ seeds.}
\label{tab:isolating_distillation}
\end{table*}

%% file: appendix.tex
\section{More Related Works}
\label{sec:more-related_work}

\subsection{Mode Connected Ensembles}

\paragraph{Population Parameter Averaging.} \cite{jolicoeur-martineau_population_2023} propose an algorithm that simultaneously trains all ensemble members. To exploit and preserve linear mode connectivity among members, they repeatedly train the members for a number of epochs on different SGD noise and data augmentations, and replace each member with the average weight vector every fifth epoch. 
\begin{figure*}
    \centering
    \includegraphics[width=\textwidth]{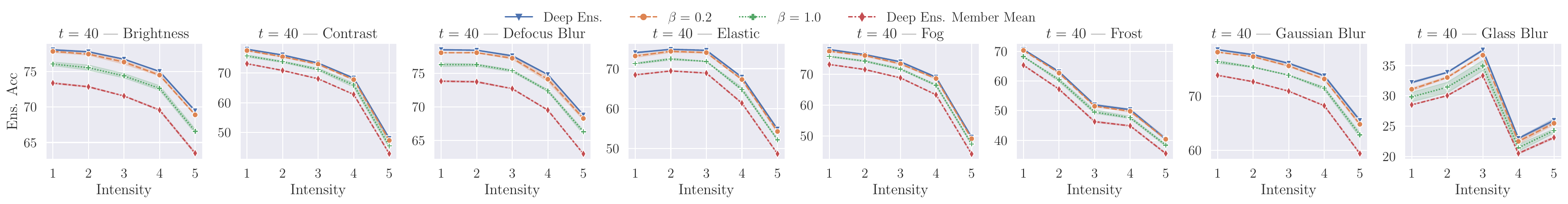}
    \caption{Out-of-distribution robustness of different ResNet20 ensembles on the corrupted CIFAR100 datasets from \cite{hendrycks_benchmarking_2019}.}
    \label{fig:ood-robustness}
\end{figure*}

\paragraph{Subspace Learning.} \cite{wortsman_learning_2021} present a method that learns subspaces of networks by uniformly sampling a network from within the subspace to be learned and backpropagating the error signal to the networks spanning the space. They prevent a collapse of the subspace by adding a penalty that encourages orthogonality between the networks constituting the subspace. In a similar manner, \citep{benton_loss_2021} present a procedure to find low-loss complexes consisting of simplexes whose vertices are found through standard training. A collapse is prevented through a regularization penalty that encourages subspaces with large volume.

\paragraph{Ensembling from a Loss Landscape Perspective.} \cite{fort_deep_2020} conclude that subspace sampling methods (weight averaging, MC dropout, local Gaussian approximations) yield solutions that lack diversity in function space and therefore, when comparing against deep ensembles, offer an inferior diversity-accuracy tradeoff. We would like to stress that this finding, which is seemingly contradictory to our work, is not applicable in our setting. The methods used in \cite{fort_deep_2020} start exploring the subspace starting from a fully optimized solution whereas our approach starts exploration right after reaching the stability point. Moreover, the methods employed by \cite{fort_deep_2020} are different to using our distillation approach, which, we argue, is more informative, but admittedly not computationally efficient.

\paragraph{Snapshot Ensembles (SSE).} \cite{huang_snapshot_2017} use a cyclic cosine annealing learning rate schedule to sample multiple models (\emph{snapshots}) within a single training run and ensemble their predictions. Their results demonstrate that snapshots acquired in later cycles become linearly mode connected as they lie in the same basin (cf. Figure 4 in \cite{huang_snapshot_2017}). Their key insight is that snapshots that lie in the same basin only add limited predictive diversity, suggesting that a cyclical learning rate schedule with a sufficiently high peak learning rate is crucial to escape the basin and attain sufficient predictive diversity for the ensemble to perform well. Our investigation makes the case that staying in the same basin and maintaining linear mode connectivity is not necessarily an impediment to achieving predictive diversity. Put differently, the benefit of ensembling does not depend on escaping the local minima the previous models are sampled from. 

\paragraph{Fast Geometric Ensembling (FGE).} Fast Geometric Ensembling (FGE) \citep{garipov_loss_2018} is similar in spirit to SSE. According to \cite{garipov_loss_2018}, the feature that distinguishes FGE from SSE is the shorter cycle length between sampling a model. FGE intentionally takes fewer steps in-between sampling models in order to not leave a region of low loss.

\paragraph{Stochastic Weight Averaging (SWA).} \cite{izmailov_averaging_2018} average weights along the SGD trajectory using a cyclical or constant learning rate. They find that averaging SGD iterates, which often lie on the edge of a loss basin, leads to a more central point and thus to an increase in generalization performance.

\paragraph{Combining SSE, FGE, and SWA.} We decided to use a procedure that combines elements from SSE, FGE, and SWA as a baseline. We argue that this approach is most effective at training an ensemble while ensuring linear mode connectivity and computational comparability, at training and inference time, with deep ensembles. As outlined in the main text, we refer to this method as Stochastic Weight Ensembling (SWE). More specifically, SWE is ensembling models in function space, acquiring them using a sequential procedure. We first decay the learning rate to a level that enables exploration of the basin without leaving it, and keep the learning rate constant thereafter. We sample a model every $T$ epochs where $T$ is on the order of epochs required to train a single model. The difference to SSE is that we specifically do \textit{not} encourage exploration of different basins and thus refrain from cyclically increasing the learning rate. The procedure is also different from SWA, as we do not average in weight space, but in function space. Lastly, it is also different from FGE, as the cycle length is comparable to that of SSE, ruling out the \textit{fast} in FGE.

\subsection{Diversity in Deep Ensembles}

As mentioned in the introduction, it is commonly believed that encouraging predictive diversity is a prerequisite for improving ensemble performance. This belief is derived from classical results in statistics on bagging and boosting weak learners \citep{freund_short_1999, breiman_bagging_1996}. While it is true that disagreement among members is a necessary condition for an ensemble to outperform any single member, recent work has shown that encouraging predictive diversity can be detrimental to the performance of deep ensembles with high-capacity members \citep{abe_pathologies_2023}. In other words, the intuition from those classical results might not be applicable. The counter-intuitive observation of \cite{abe_pathologies_2023} is explained by the fact that diversity encouraging penalties affect all predictions irrespective of their correctness. As a result, these penalties can adversely affect the performance of individual members, which in turn can undermine the performance of the ensemble. 

\section{More Experiments}

\begin{figure}
    \centering
    \includegraphics[width=\linewidth]{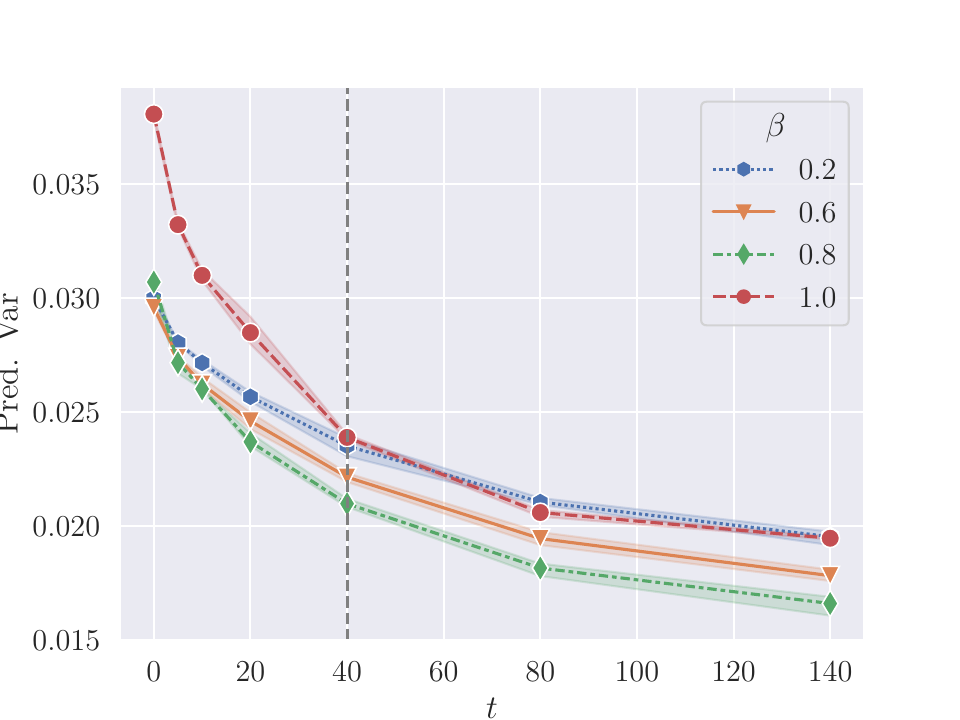}
    \caption{Predictive variance as a function of time parameter $t$ for ResNet20 ensembles on CIFAR100. The dashed vertical lines mark the $t$ used in Table~\ref{tab:beta_table}.}
    \label{fig:t-ablation-pred-var}
\end{figure}

\paragraph{Predictive Variance.} 
In addition to the Jensen-Shannon divergence in Fig.~\ref{fig:t-ablation-jsd}, we also analyzed the predictive variance. We display the results in Fig.~\ref{fig:t-ablation-pred-var} and observe that the results corroborate the conclusions drawn in the main text. 

\paragraph{Out-of-distribution Robustness.} A significant advantage of deep ensembles over single models is their robustness towards distribution shifts \citep{lakshminarayanan_simple_2017}. Thus, we evaluate whether connected ensembles can match the out-of-distribution robustness of multi-basin deep ensembles in Fig.~\ref{fig:ood-robustness}. We use a subset of corruptions from CIFAR100-C \citep{hendrycks_benchmarking_2019} and note that distilled ensembles exhibit a similar degree of robustness to distribution shifts as their multi-basin counterparts.

\section{Implementation Details}
\label{sec:implementation_details}

\begin{table*}
\resizebox{1.0\linewidth}{!}{
\begin{tabular}{@{}llccccccccccccc@{}}\toprule
&& \multicolumn{1}{c}{Deep Ens.} & \phantom{abc}& \multicolumn{1}{c}{SWE} & \phantom{abc}& \multicolumn{4}{c}{Distilled Ens.}  & \phantom{abc} & \multicolumn{4}{c}{Constrained Ens.} \\
 \cmidrule{7-10} \cmidrule{12-15}
&& $T$ && $T$ && $\beta$ & $T$ & $t$ & Dist. Epochs && $\beta$ & $T$ & $t$ & Dist. Epochs\\

\midrule
\multirow{2}{*}{CIFAR10}

& ResNet20  & $110$ && $110$ && $0.2$ & $110$ & $10$ & $100$ && $1.0$ & $110$ & $10$ & $100$ \\
& ViT       & $165$ && ----  && $0.2$ & $165$ & $15$ & $150$ && $1.0$ & $165$ & $15$ & $150$\\

\midrule
\multirow{2}{*}{CIFAR100}

& ResNet20  & $190$ && $190$ && $0.2$ & $190$ & $40$ & $150$ && $1.0$ & $190$ & $40$ & $150$ \\
& ViT       & $165$ && ----  && $0.2$ & $165$ & $15$ & $150$ && $1.0$ & $165$ & $15$ & $150$ \\

\midrule
\multirow{2}{*}{Tiny ImageNet}

& ResNet20  & $130$ && $130$ && $0.2$ & $130$ & $30$ & $100$ && $1.0$ & $130$ & $30$ & $100$ \\
& ViT       & $140$ && ----  && $0.2$ & $140$ & $15$ & $125$ && $1.0$ & $140$ & $15$ & $125$ \\

\bottomrule
\end{tabular}
\caption{Computational cost of ensembles. For deep ensembles, $T$ refers to the number of epochs per sample. Similarly, for SWE, $T$ is the cycle length in-between taking a sample. For constrained and distilled ensembles, $t$ is the epoch after which we split the runs and starting distilling for $\text{Dist. Epochs}$.}
\label{tab:computational_cost}
}
\end{table*}
\paragraph{Computational Cost.} If not stated otherwise, we consider ensembles of size $M=5$. Table~\ref{tab:computational_cost} illustrates the computational cost on a per model basis.

\paragraph{Optimizers.} With the exception of experiments conducted with ViTs, we use SGD as an optimizer with a peak learning rate of $0.1$. We use a cosine decay schedule with linear warmup for the first $10\%$ of training. Momentum is set to $0.9$. For ViTs, we use Adam \citep{kingma_adam_2015} with $\beta_1=0.9$ and $\beta_2=0.999$. The batch size is at 128 and we set the temperature in the distillation experiments to $\tau=3$. For SWE, we apply the same linear warmup cosine decay schedule as for the other ensemble methods, but stop decaying the learning rate at 0.01 and hold it constant thereafter to enable exploration of the basin.

\paragraph{Datasets.} We experiment with the classic image classification baselines CIFAR \citep{krizhevsky_learning_2009} and Tiny ImageNet \citep{le_tiny_2015}. For all experiments, we make use of data augmentation. More specifically, we use horizontal flips, random crops, and color jittering.

\paragraph{Architectures.} We use the ResNet20 implementation from \cite{ainsworth_git_2023} with three blocks of 64, 128, and 256 channels, respectively. We note that this implementation uses LayerNorm \citep{ba_layer_2016} instead of BatchNorm \citep{ioffe_batch_2015}, as it eliminates the burden of recalibrating the BatchNorm statistics when interpolating between networks. Our Vision Transformer implementation is based on \cite{lippe_uva_2022} and composed of six attention layers with eight heads, latent vector size of 256 and hidden dimensionality of 512. We apply it to flattened $4\times4$ image patches.

\paragraph{Permuted Ensembles.} We use the \textsc{PermutationCoordinateDescent} implementation from \cite{ainsworth_git_2023} to bring deep ensemble models into alignment. The implementation of the \textsc{PermutationCoordinateDescent} algorithm can be found at \url{https://github.com/samuela/git-re-basin}.

\paragraph{Joint Connectivity.} As mentioned in the main text, we draw samples $\bm{\lambda}_1, \dots, \bm{\lambda}_N \sim \text{Dir}(\bm{1})$ to approximately assess the joint connectivity of ensemble members. For each seed, we evaluate $N=50$ samples and compute $\bar{q}_{\text{joint}} = \frac{1}{N}\sum_{i=1}^N q_{\text{joint}}(\bm{\lambda}_{i})$

\paragraph{Hardware.} We ran experiments on a cluster with NVIDIA GeForce RTX 2080 Ti and NVIDIA GeForce RTX 3090 GPUs.